\documentclass[letterpaper]{article} 
\usepackage[draft]{aaai2027}
\usepackage[hyphens]{url}  
\usepackage{graphicx} 
\usepackage{natbib}  
\usepackage{caption} 
\usepackage{algorithm}
\usepackage{algorithmic}
\usepackage{amsmath}

\usepackage{hyperref}
\makeatletter
\renewcommand{\@oddhead}{}
\renewcommand{\@evenhead}{}
\renewcommand{\@oddfoot}{\hfil\thepage\hfil}
\renewcommand{\@evenfoot}{\hfil\thepage\hfil}
\makeatother

\usepackage{newfloat}
\usepackage{listings}
\DeclareCaptionStyle{ruled}{labelfont=normalfont,labelsep=colon,strut=off} 
\floatstyle{ruled}
\newfloat{listing}{tb}{lst}{}
\floatname{listing}{Listing}

\usepackage{booktabs}

\title{Autonomous Research for Open-Ended Problems:\\ A Case Study on Telecom Ticket Retrieval}
\author {
    Junghyun Min\textsuperscript{\rm 1\rm *} \quad
    Huseyin Uzunalioglu\textsuperscript{\rm 2} \quad
    Mohamed Trabelsi\textsuperscript{\rm 2}
}
\affiliations {
    \textsuperscript{\rm 1}Georgetown University\\
    \textsuperscript{\rm 2}Nokia Bell Labs\\
    \texttt{jm3743@georgetown.edu, \{huseyin.uzunalioglu, mohamed.trabelsi\}@nokia-bell-labs.com} \\
    \small\textsuperscript{*}Work performed as an intern at Nokia Bell Labs
}

\begin{document}

\maketitle
\thispagestyle{plain}

\begin{abstract}
Recent breakthroughs in LLM-based systems and their abilities in problem solving and coding have allowed progress in the AI for Science paradigm, potentially replacing human roles in machine learning (ML) research.
However, while several frameworks of fully autonomous end-to-end ML research have been proposed, successful implementations of them are often limited to problems with narrow search spaces, like language modeling or biomedical ML benchmarks.
In this paper, we explore how autonomous research can be adapted to solve open-ended, industry-grade ML problems, by considering a case study: telecom ticket retrieval, an open-ended task with degrees of freedom in representation, architecture, and training data generation.
We discover that autonomous research for open-ended problems with commercial and open-source agents shows both promise and limitations: 
while autonomous research can excel in narrow hyperparameter optimization, it lacks human-like intuition and creativity and requires operational overhead.
Even with minimal human supervision, autonomous research can reach $90\%$ of state-of-the-art performance (0.34 vs. 0.38 Recall@1) in a much shorter time period (10 weeks vs. 10 months of human work) at a modest cost (up to \$200 per Cursor campaign).
Our empirical evidence recommends that human researchers and autonomous research frameworks work together for best results in ML research.
\end{abstract}


\section{Introduction}
Autonomous research is a subset of the broader AI for Science paradigm, where autonomous agents replace human decision-making in research.
The rise in utility of LLMs and their derivative reasoning \citep[e.g.][]{openai_openai_2026} and coding agents \citep[e.g.][]{research_composer_2026} have led to implementations of autonomous research, partially replacing human elements in research \citep[e.g. in hypothesis generation or in experiment implementation;][]{xu_artificial_2021, van_noorden_ai_2023} or fully automating research end-to-end \citep{lu_ai_2024, yamada_ai_2025, lu_towards_2026}.

Within autonomous ML research, applications of autonomous research for ML problems, a minimalistic implementation that iteratively proposes new experiments to hypothesize and verify improvement against a pre-determined performance metric (e.g. language modeling loss) popularized the approach \citep{karpathy_karpathyautoresearch_2026}.
In each of its iterations, an LLM-powered agent proposes a new ML system design that could improve on the metric (i.e. lower the loss), implements and runs the experiment, before evaluating the system and verifying the improvement.
The repository has spurred a series of extensions, including a multi-agent adaptation that bases experiment proposals on inter-agent discussions \citep{gao_autoscientists_2026} and a human-in-the-loop adaptation focusing on AI-human collaboration \citep{liu_autoresearchclaw_2026}.
However, most successful implementations have been limited to simpler tasks with fixed training data and a small search space like language modeling or biomedical ML benchmarks.
In this work, we ask: \textbf{how, if possible, can we apply end-to-end autonomous research to open-ended tasks?}

This question is broad; we thus break it down into three research questions.
\begin{itemize}
    \item \textbf{RQ1}: How does autonomous research output compare to output from human research in terms of downstream open-ended task performance?
    \item \textbf{RQ2}: What is the cost of performing autonomous research for open-ended problems in time and API charges?
    \item \textbf{RQ3}: What are empirical advantages and limitations of applying autonomous research to open-ended problems?
\end{itemize}


To investigate RQs 1-3, we perform a case study on a prime example of an open-ended task: \textbf{telecom ticket retrieval}, whose degrees of freedom span not only over architecture and hyperparameter optimization, but also training data generation and structured document representation (Section \ref{sec:telecom-ticket-retrieval}).
The open-ended nature of the task warrants additional harness for reliable autonomous research campaigns, which we describe in Section \ref{sec:additional-harness}.

We also aim to understand how agent design choices affect answers to the research questions.
Thus, we investigate the three questions by systematically experimenting with varying implementations of agentic autonomous research pipelines across three design choices: how many agents to spawn for the research campaign, which underlying LLM agent to employ, and whether to inform the agent about human-designed solutions (Section \ref{sec:axes}).

We find that the best discovered model is a single fine-tuned retriever that fails to outperform the internal state-of-the-art model that incorporates ``creative'' solutions like model ensembling, data augmentation, and re-ranking, but substantially outperforms the human-designed fine-tuned retrieval model (Section \ref{sec:discovered_pipeline}).
In addition, perhaps surprisingly, we find that the framework structure involved in the campaigns, the underlying LLM agent, or the informedness do \textbf{not} seriously affect downstream task performance, with a locally hosted open-weight GPT-OSS with no prior knowledge outperforming a team of Claude Sonnet 5 agents with prior knowledge (Section \ref{sec:selection-effects}, Table \ref{tab:campaigns}).

Overall, our empirical evidence shows that today's available commercial and open-weight agents are great at deep, narrow hyperparameter sweeps yet lack human-like intuition or creativity.
Thus, we recommend the following takeaways (Section \ref{sec:insights}: (1) autonomous research is best complemented by human ``directing''; (2) autonomous research saves time for a modest cost; and (3) operational struggles that require human ``babysitting'' exist.

\section{Background and Related Work}
\label{sec:background}
In this work, we investigate the generalizability of autonomous research to open-ended tasks.
We opt to task our implementations of autonomous research (detailed in Section \ref{sec:additional-harness}) with solving telecom ticket retrieval (TTR).
We use this section to describe the background of this work in autonomous research (Section \ref{sec:system}), and in TTR (Section  \ref{sec:telecom-ticket-retrieval}).

\subsection{Autonomous Research}
\label{sec:system}

Our work builds on previous work towards AI for Science and autonomous research.
Recent advances in LLMs, reasoning and coding agents \citep{openai_openai_2026, research_composer_2026} have presented a possibility to automate the scientific method.
\citet{xu_artificial_2021} discuss the potential of using artificial intelligence as a tool to accelerate scientific research across applied and fundamental sciences.
While early applications were limited to hypothesis generation or experimental implementation \citep[e.g.][]{van_noorden_ai_2023}, \citet{lu_ai_2024} proposed the first end-to-end ``AI Scientist'' that generates novel research ideas, performs experiments, and even writes a paper based on the results.
This full automation of the scientific method, while still limited in scope, quality, and reliability, has only further developed since then
 \citep{yamada_ai_2025, lu_towards_2026}.
For a comprehensive review on autonomous research, we refer readers to \citet{tie_autoresearch_2026}.


Our work utilizes a single-agent and a multi-agent implementations, based on two recent frameworks of autonomous research, namely \texttt{autoresearch} \citep{karpathy_karpathyautoresearch_2026} and \texttt{AutoScientists} \citep{gao_autoscientists_2026}.

\paragraph{Single-agent \texttt{autoresearch}.}
\citeauthor{karpathy_karpathyautoresearch_2026}'s framework is a single-agent loop designed for code-level optimization within a single Python script. 
The agent, a command-line interface tool with an underlying LLM (e.g. Claude Code, Cursor, and Pi), operates through an iterative process: it reads the repository state, modifies training or architecture parameters, runs trials, and evaluates performance. 
If the target metric improves, the change is committed to version control; if the metric degrades or trial execution fails, the repository is reverted to the prior state.

\paragraph{Multi-agent \texttt{AutoScientists}.}
In contrast, \citet{gao_autoscientists_2026} implement \texttt{AutoScientists}, a multi-agent framework that divides the research workflow across specializing sub-agents.
A central orchestrator agent, also a command-line interface tool with an underlying LLM, manages sub-agents, each of which is assigned to one sub-task e.g. hypothesis generation, code modification, and data analysis.
Sub-agents communicate via a messaging board, implemented via a ClawInstitute API\footnote{\url{https://clawinstitute.aiscientist.tools/}}, discussing previous experimental results and proposing future experiment direction.

\subsection{Telecom Ticket Retrieval}
\label{sec:telecom-ticket-retrieval}

Our case study concerns telecom ticket retrieval, a document retrieval task in the telecommunications domain.
In large-scale industrial systems, problematic incidents are often flagged by tickets, which include the details of and the context around the incidents.
Directing, processing, and resolving such tickets are core difficulties in system operations that often require domain expertise, institutional knowledge, and human and financial resources \citep{fuchs_improving_2022}.
Attempts to automate ticket directing or resolution have benefitted from the recent advancements in natural language processing technologies \citep{dasgupta_towards_2014, zhou_star_2017, ali_zaidi_multiapproach_2022, marcuzzo_multi-level_2022, jain_ai_2025}, including neural representation learners \citep[e.g.][]{devlin_bert_2019, reimers_sentence-bert_2019} and generative large language models \citep[LLMs; e.g.][]{openai_gpt-oss-120b_2025, xu_qwen3-omni_2025}.

Telecommunications network operation is no exception to this trend, with recent work proposing deep neural representations for abstractive document summarization \citep{trabelsi_absformer_2023} and effective solution recommendation \citep{saragadam_scalable_2025}; end-to-end multi-agent systems that detect and resolve problematic incidents \citep{shi_leveraging_2026}; and generative LLMs and agentic systems that retrieve similar documents to incident tickets and automatically generate resolution documents \citep{trabelsi_teledoctr_2026, javaji_astra_2026}.

Telecom ticket retrieval is a subset of this wider attempt to automate ticket resolution in the telecommunications domain, and is the task of retrieving a relevant resolution document given a query document.
We define the task as follows:
Let $D_{\text{raw}} = \{(q_i, t_{i}^*)\}_{i=1}^N$ denote a domain dataset comprising $N$ operational incident queries $q_i \in \mathcal{Q}$ and their corresponding ground-truth resolution documents $t_{i}^* \in \mathcal{T}_{\text{raw}}$\footnote{There may be more than one ground-truth resolution documents tied to a ticket, as a single incident may be described across multiple tickets and resolved across multiple resolution documents.}. 

We define a document preprocessing transformation $\mathcal{P}: \mathcal{T}_{\text{raw}} \to \mathcal{T}$ that cleans, canonicalizes, or otherwise pre-processes raw documents. Let $E_\theta(\cdot)$ denote a dense text encoder parameterized by weights $\theta$. The retrieval relevance score between a query $q$ and a preprocessed ticket $t = \mathcal{P}(t_{\text{raw}})$ is computed via vector cosine similarity:

$$S_\theta(q, t) = \frac{E_\theta(q) \cdot E_\theta(\mathcal{P}(t_{\text{raw}}))}{\|E_\theta(q)\|_2 \, \|E_\theta(\mathcal{P}(t_{\text{raw}}))\|_2}$$

The objective of an autonomous research agent would be to jointly discover the optimal preprocessing function $\mathcal{P}^*$, sampling hyperparameter configuration $\lambda^*$, and encoder weights $\theta^*$ that maximize top-$K$ recall over the domain evaluation split:

$$\mathcal{P}^*, \lambda^*, \theta^* = \arg\max_{\mathcal{P}, \lambda, \theta} \frac{1}{N} \\ \sum_{i=1}^N \mathbf{1}\left( t_i^* \in K_{t \in \mathcal{T}} \left( S_\theta(q_i, \mathcal{P}(t_{\text{raw}})) \right) \right)$$

where $\mathbf{1}(\cdot)$ is the indicator function evaluating whether the target document falls within the top-$K$ retrieved candidates.
The wide search space across $\mathcal{P}$, $\lambda$, and $\theta$ differentiates telecom ticket retrieval from other tasks that previous autonomous research implementations explore.

Overall, we believe that telecom ticket retrieval is a good testbed for applying autonomous research for open-ended problems due to its three features: difficulty, openness, and complexity of the state-of-the-art system.
The task is \textbf{difficult} because the dataset contains highly contextual language as submitted tickets often do not contain the full details relevant to their causal incidents, consist of specific jargon that requires domain-specific understanding \citep{gururangan_dont_2020} and are highly variant in their ``contentfulness'' depending on the engineer that submitted them.
It is \textbf{open} because the degrees of freedom lie beyond just hyperparameter search; training data generation (e.g. negative sample mining), training loss design, and document representation are all part of the search space.
Finally, the human-developed state-of-the-art system based on \citet{trabelsi_teledoctr_2026} is \textbf{complex}, incorporating data augmentation, model ensembling, and re-ranking.



\section{Experimental Setup}
\label{sec:experimental_setup}

\subsection{Datasets and Evaluation}
\label{sec:data}
We work with a dataset of real-world telecom troubleshooting tickets, comprising of $250k$ incident reports (tickets: \texttt{T}) and their resolutions ($204k$ fault analyses: \texttt{FA} and $89k$ technical analyses: \texttt{TA}).
A \texttt{TA} details an intermediate investigation to a single \texttt{T}, while a \texttt{FA} is a high-level overview for a problem that spans multiple \texttt{T}s.
This way, we build a cluster of documents addressing the same problem, with each cluster possibly containing one \texttt{FA} and multiple \texttt{T}s and \texttt{TA}s.
This allows us to explore various strategies for positive pair sampling.

Evaluation is parallel to \citet{trabelsi_teledoctr_2026}, with $7.6k$ held-out queries and $1.2k$ held-out gold resolution documents, which compete against the entire dataset during evaluation.
A correct retrieval is when a query and the retrieved document are in the same cluster.
That is, they address the same incident (e.g. \texttt{T} and its resolving \texttt{FA}), or are solved by the same resolution document (e.g. \texttt{TA} of a \texttt{T$_i$} and another \texttt{TA} of \texttt{T$_j$}, where \texttt{T$_i$} and \texttt{T$_j$} are associated via a single \texttt{FA}).

Following previous work, we report retrieval performance primarily as recall@$k$ (the rate of relevant documents within top $k$ retrieved documents).

We compare our metric to baseline BM25 \citep{robertson_okapi_1994}, an MPNet-based \citep{song_mpnet_2020} fine-tuned retrieval model, an ensemble of such fine-tuned models \citep{trabelsi_teledoctr_2026}, and an unpublished internal state-of-the-art model that employs LLM-based data augmentation and re-ranking.
Our comparison models are all designed and trained by humans.

All models are trained on 4xNVIDIA RTX A6000 servers, each of whose GPUs have 48GB of VRAM.

\subsection{Harness for Operational Stability}
\label{sec:additional-harness}

To adapt autonomous research to an open-ended retrieval problem, we introduce an operational harness targeting task and environment documentation, search space definition, and experimental loop control.
These harness components are a result of many trial-and-error loops, and show varying effectiveness.
Nonetheless, we believe they constitute the most cost-effective methods of controlling the operational stability of today's available agents and autonomous research frameworks.

\paragraph{Task and environment documentation.}
Unlike language modeling where data is unstructured and single-stream \citep{radford_improving_2018}, TTR involves complex, heterogeneous document structures. 
We explicitly document defining data structure, task scope, agent role, and the primary optimization metric: $\text{Recall}@1$.
We also document available hardware resources and instruct agents to maximize GPU utilization.
Still, agents frequently underutilize compute resources.
Furthermore, GPT-OSS 120B fails to reliably orchestrate sub-agent spawning, even with access to explicit instructions\footnote{We additionally attempted campaigns with Qwen Coder 30B \citep{yang_qwen3_2025}, all of which failed to reliably spawn sub-agents.}, succeeding only once across dozens of attempts despite using a sub-agent extension\footnote{\url{https://github.com/nicobailon/pi-subagents}}.
Commercial models (Claude Sonnet 5, Cursor Composer 2.5) reliably spawn sub-agents without additional skills, extensions, or harness.

\paragraph{Search space definition.}
Without explicit guidance, autonomous research agents default to optimizing standard hyperparameters such as learning rate and number of training epochs.
To force broader exploration, we explicitly define the search space.

In the multi-agent implementation, we assign a search space to each of three agent ``teams'': representation, training data generation, and architecture \& hyperparameters.
This way, proposed experiments also explore new document representation and data generation strategies, rather than only focusing on optimizing model hyperparameters.

In the single-agent implementation, we initialize a set of variables, each of which maps to a decision in representation and dataset generation within the codebase to encourage experiments that explore beyond standard hyperparameters.
While this variable abstraction prevents focusing only on standard hyperparameters, it introduces a trade-off: agents primarily toggle pre-defined candidate options rather than implementing novel code modifications.

We note that while we believe this explicit space definition may be able to mitigate the agent's inability to reliably incorporate documented human-designed SoTA techniques (Section \ref{Sec:operation-decisions-results}), we did not include such implementation as we believe it would defeat the purpose of using autonomous research to ``discover'' a new system for our task.

\paragraph{Scripted loop.}
Prompting agents to manage their own execution loops (e.g., instructing them to ``never stop'') or implementing Python-based heartbeat controllers proved unreliable, frequently resulting in stalled agents or redundant experiments. 
We replace agentic self-looping with a deterministic outer bash script. 
The harness initializes an agent session, then immediately exits.
Upon exiting, a session resume key can be retrieved, and a bash while loop can be executed:
\begin{verbatim}
while true; do
    pi --resume <SESSION_KEY> -p
        "Continue experimenting to
        improve Recall@1. Feel free
        to refer to program.md."
done
\end{verbatim}
We find this is the most effective implementation of a never-stopping loop that ensures continuity in the autonomous research campaigns until the human researcher stops the loop.

\begin{table*}[h]
    \small
    \centering
        \caption{A table of experimental settings across variable axes. Each experiment is associated with one variable from each axis.}
    \label{tab:ablations}
    \begin{tabular}{r|l}
    \toprule
        \textbf{Axis} & \textbf{Variables} \\
        \midrule
        Framework structure & Single-agent loop, multi-agent team \\
        Agent choice & Claude Sonnet 5, Cursor Composer 2.5, GPT-OSS 120B\\
        Informedness & Uninformed, SoTA components provided in documentation (informed) \\
        \bottomrule
    \end{tabular}

\end{table*}

\subsection{Autonomous Research Implementations}
\label{sec:axes}
To understand how implementation design choices affect autonomous research performance, we analyze downstream task performance across three axes of design choices: across framework structure, LLM agent choice, and informedness.
We describe our variables across the three axes in Table \ref{tab:ablations}; each autonomous research campaign is associated with one variable from each axis.
For example, one campaign may use a single GPT-OSS-based agent with no information about existing SoTA provided in the documentation.
Another may use multiple with Claude Sonnet-based agents with knowledge of components of the existing SoTA provided in the documentation.

\paragraph{Across framework structures.}
We evaluate two framework structures: a single-agent loop extended from \citet{karpathy_karpathyautoresearch_2026} and a multi-agent team extended from \citet{gao_autoscientists_2026}.
We hypothesize that the multi-agent implementation's stronger performance than the single-agent implementation in tasks with narrower search spaces \citep{gao_autoscientists_2026} will generalize to open-ended tasks like ticket retrieval.
It is also intuitive to predict that a team of agents will outperform a single agent in autonomous research.

\paragraph{Across LLM agents.}
We compare performance across two commercial agents--Claude Sonnet 5 \citep{anthropic_claude_2026} and Cursor Composer 2.5 \citep{research_composer_2026}--and one open-weight model, GPT-OSS 120B \citep{openai_gpt-oss-120b_2025} hosted on Ollama \footnote{\url{https://ollama.com/}} as a backend of an open-source coding agent Pi\footnote{\url{https://pi.dev}}.
We hypothesize that using Claude as the underlying agent for our autonomous research campaign will outperform using Cursor; and in turn, that Cursor will outperform GPT-OSS, based on the reported difference in reasoning abilities of them and their related models \citep{yueh-han_reasoning_2026, metr_details_2024, huang_olympicarena_2024}.

\paragraph{Across informedness.}
Finally, we compare an uninformed start from scratch against an ``informed'' setup where system documentation explicitly informs the agent of human-designed SoTA techniques, including generative data augmentation, re-ranking, and model ensembling.
Here, we hypothesize that providing information about the components of the existing SoTA system will result in higher task performance than not providing any information.


\section{Results}
\subsection{Our Best Discovered Model}
\label{sec:discovered_pipeline}
While we run multiple autonomous research campaigns as we describe in Section \ref{sec:axes}, we report our best discovered system.
As seen in Table \ref{tab:comparison-results}, the resulting system is a single fine-tuned MPNet retriever.
Interestingly, the single-model system outperforms an ensemble system, and even nears the performance of the internal SoTA system, with $\text{Recall}@1$ of \textbf{0.343}, $\text{Recall}@10 = \mathbf{0.630}$, $\text{Recall}@50 = \mathbf{0.768}$.
We detail the system and its document representation and pre-processing, training data generation, and architecture and hyperparameters in Appendix \ref{sec:details-best-model}.

\begin{table*}[]
    \small
    \centering
    \caption{Our single fine-tuned model against comparison systems.}
     \label{tab:comparison-results}
    \begin{tabular}{l|cccc}
    \toprule
     \textbf{Model} & \textbf{Recall@1} & \textbf{Recall@10} & \textbf{Recall@50} \\
     \midrule
     BM25 \citep{robertson_okapi_1994} & 0.072 & 0.113 & 0.234 \\
     Fine-tuned MPNet \citep{song_mpnet_2020} & 0.200 & 0.400 & 0.571 \\
     Ensemble \citep{trabelsi_teledoctr_2026} & 0.251 & 0.461 & 0.627 \\
     Internal SoTA & \textbf{0.380} & \textbf{0.783} & - \\
     \midrule
     Discovered fine-tuned MPNet (ours) & 0.343 & 0.630 & 0.768 \\
     \bottomrule
\end{tabular}

\end{table*}

\subsection{Answering RQs}
\label{sec:selection-effects}

Across implementations, varying the number and type of agents and the documentation they had access to had negligible impact on peak task performance.
The best model (Section \ref{sec:discovered_pipeline}) was produced by the single-agent Cursor loop informed of knowledge of human-designed SoTA\footnote{Although, the agent did not make use of this information.} after 17 unique experiments.
We report a summary of our findings across in Table \ref{tab:campaigns} and in paragraphs below.

\begin{table*}[htbp]
\small
\centering
    \caption{Comparison of autonomous research campaigns across framework structure (single- and multi-agent), agent (Claude, Cursor, and GPT-OSS), and informedness (informed and uninformed).}
\label{tab:campaigns}
\begin{tabular}{l l l c c p{8cm}}
\toprule
\textbf{\# agents} & \textbf{Agent} & \textbf{Inf.} & \textbf{\# Exps.} & \textbf{Best R@1} & \textbf{Findings from Best Run} \\
\midrule
Single & Cursor & Uninf. & 33 & 0.2744 & Added \texttt{T}-\texttt{TA} pair types, trained for 14 epochs with batch size 64, and scaled training to the full corpus. \\
Single & Cursor & Inf. & 17 & \textbf{0.3433} & Applied $2\times$ oversampling to peer ticket (\texttt{T}- \texttt{T}) pairs and expanded encoder sequence length to 384 tokens. \\
Single & Claude & Uninf. & 9 & 0.3378 & Expanded pair types (\texttt{T}-\texttt{FA}, \texttt{FA}-\texttt{T}, \texttt{T}-\texttt{TA}, \texttt{TA}- \texttt{T}) and set context length to 384 tokens. \\
Single & GPT-OSS & Uninf. & 12 & 0.3353 & Applied full pair-type oversampling (\texttt{T}-\texttt{FA}:3, \texttt{FA}-\texttt{T}:3, \texttt{T}-\texttt{TA}:2, \texttt{T}-\texttt{T}:2) and raised sequence length to 320 tokens. \\
Single & GPT-OSS & Inf. & 13 & 0.3244 & Introduced $\texttt{FA}- \texttt{TA}$ pair types with 2 training epochs, batch size 64, learning rate $3\times 10^{-5}$, and 4 hard negatives per pair. \\
Multi & GPT-OSS & Uninf. & 4 & 0.2043 & Fine-tuned MiniLM-L6-v2 baseline trained with \texttt{T}-\texttt{FA} pairs. \\
Multi & Cursor & Uninf. & 132 & 0.2652 & Set learning rate to $7.5\times 10^{-5}$, anchor pair cap to 11, and expanded sequence length to 384 tokens. \\
Multi & Claude & Inf. & 18 & 0.2400 & Mined embedding-similarity hard negatives (ranks 3–20) using a frozen base encoder. \\

\bottomrule
\end{tabular}
\end{table*}

\paragraph{How does autonomous research output compare to output from human research in terms of task performance?}
As shown in Table~\ref{tab:comparison-results}, our best system ($\text{Recall}@1 = 0.343$, $\text{Recall}@10 = 0.630$) substantially outperforms a BM25 baseline ($\text{Recall}@1 = 0.072$), standard fine-tuned model (MPNet; $\text{Recall}@1 = 0.200$), and human-designed ensembles \citep[][$\text{Recall}@1 = 0.251$]{trabelsi_teledoctr_2026}.
However, this system does not surpass the human-designed Internal SoTA ($\text{Recall}@1 = 0.380$, $\text{Recall}@10 = 0.783$), which incorporates structural components not attempted by the agent like re-ranking, data augmentation, and model ensembling.

\paragraph{What is the cost of performing autonomous research in time and API charges?}
Conducting a single 10-20 experiment research campaign using Cursor Composer 2.5 costs an estimated \$150–\$200 in API charges, excluding the initial cost for repository setup and harness development. 
Equivalent runs with Claude Sonnet 5 cost approximately 2–3$\times$ more (\$300–\$600), while open-weight models (e.g. GPT-OSS) can be hosted locally with no API or subscription cost. 
In terms of development time, the 17-run autonomous campaign completed within 1 GPU-week of compute time and 10 weeks of end-to-end execution.
In contrast, human-engineered comparison systems required 10 months of development.

\paragraph{What are empirical advantages and limitations of applying autonomous research to open-ended problems?}
The primary advantage of applying autonomous research is its ability to perform a focused optimization within a narrowly fixed search space.
However, agents exhibit a fundamental lack of intuition-driven hypothesis generation. 
Rather than considering training dynamics (e.g. loss trajectory analysis, early stopping) or training data distribution (e.g. average or median length), the agent arbitrarily changed parameters (e.g., lengthening training from 1 to 2 epochs; increasing sequence length from 256 to 384).
In Section \ref{sec:insights}, we discuss empirical takeaways in greater detail.

\subsection{Effects of Operational Decisions}
\label{Sec:operation-decisions-results}
In Section \ref{sec:axes} and Table \ref{tab:ablations}, we describe our various implementations of autonomous research.
We summarize our takeaways from campaigns across framework structure, LLM agent choice, and informedness in this section.

\paragraph{Across framework structures.}
We initially hypothesized that discussion among agents would allow the multi-agent implementations to produce higher performance systems via wider searches and stronger experiment proposals.
Surprisingly, however, the framework structure involved had no substantial impact on peak downstream performance.
In fact, the best model was produced by a single-agent loop.
While multi-agent teams distributed search spaces, their discussions rarely introduced effective experimental proposals beyond parameter tuning.

\paragraph{Across LLM agents.}
Following LLM benchmarking work \citep{metr_details_2024, huang_olympicarena_2024, yueh-han_reasoning_2026}, we expected reasoning ability (Sonnet 5 > Composer 2.5 > GPT-OSS 120B) to correlate with downstream task performance.
However, our results demonstrate that LLM reasoning does not substantially affect downstream performance.
Half of our campaigns converged to a tight performance band, with Recall@1 between 0.30 and 0.34. While Cursor Composer 2.5 generated the top-performing single model ($\text{Recall}@1 = 0.343$), no single agent produced a particularly strong downstream task performance.
In our campaigns, agent choice does not appear to substantially affect the upper bound of discovered system performance.

\paragraph{Across informedness.}
Finally, we expected informed campaigns to outperform uninformed campaigns.
Counterintuitively again, this ablation had little effect on downstream task performance. 
This is not to say information on successful systems or human intuition about what should work is not helpful; instead, this is due to agents failing to consider these suggestions.
Regardless of explicit documentation, no agent attempted to implement re-ranking, data augmentation, or model ensembling, consistently defaulting to single-model fine-tuning.
This demonstrates a fundamental limitation in current agent behavior: providing human insight in the form of documentation is insufficient to induce ``outside-the-box'' architectural changes.

\section{Insights}
\label{sec:insights}
In this work, we evaluate the generalizability of autonomous research to open-ended machine learning tasks by conducting a case study on telecom ticket retrieval.
Our findings yield three primary takeaways regarding performance boundaries, engineering efficiency, and operational bottlenecks.

\paragraph{Autonomous research and human intuition can complement each other in ML research.}
The agent-discovered pipelines are all single fine-tuned systems.
Still, they surpass human-designed single-model baselines and ensembles, although they underperform complex state-of-the-art systems.
Agents search for optimal hyperparameter settings and data representation strategies, but fail to expand their search to high level, systemic strategies like re-ranking, synthetic data augmentation, or model ensembling.
They were also conservative in exploring alternate representation encoders.
While single-agent campaigns started with MPNet \citep{song_mpnet_2020} and multi-agent runs with SentenceBERT's MiniLM \citep{reimers_sentence-bert_2019, wang_minilmv2_2021}, both rarely proposed moving towards a more modern or larger architecture.
The only other proposed encoder was BAAI General Embedding model \citep{chen_m3-embedding_2024}.

In addition, agents are susceptible to suboptimal follow-up proposals, e.g. proposing to extend training from 1 to 2 epochs rather than proposing a long training run with early stopping or analyzing the loss curve, or proposing to extend maximum document length from 256 to an arbitrary 384 tokens, instead of analyzing the length distribution of the documents in the dataset.

This pattern reflects documented limits in LLM reasoning \citep{zahavy_position_2026, balani_llms_2026} and the structural constraints imposed by next token-based pre-training objectives \citep{mccoy_embers_2024}. 
Effective deployment still requires human researchers to define the search space and select implementation specifications, relying on their expertise, domain knowledge, intuition and reasoning to guide and successfully complete an autonomous research campaign.

In addition, given highly optimized single-model systems from autonomous research, even when autonomous research agents struggle to propose ``outside-the-box'' solutions, combining such a system with human-proposed solutions like ensembling, re-ranking, data augmentation could be an effective way to maximize utility from autonomous research.
That is, autonomous research and human intuition can complement each other in ML research.

\paragraph{Autonomous research saves time for a modest cost.}
In our case study, autonomous research was able to substantially reduce overall development time. 
Our work was designed and completed in 10 weeks, compared to around 10 months of human engineering required to build internal state-of-the-art baselines.
Of course, this cannot be a direct comparison since the human-led and agentic research efforts are not independent of each other.

In terms of monetary cost, while an exact monetary cost has not been provided due to the structure of the group's billing system, we estimate that a single campaign has a cost around US\$150-200 on Cursor, not including the amount that was used to fine-tune documentation and the harness.
We estimate the cost to be greater by a factor of 2-3 on Claude, and the cost is negligible on GPT-OSS, given that GPU resources are already available.
Considering GPT-OSS performs similarly to its commercial counterparts (Section \ref{Sec:operation-decisions-results}), autonomous research can help accelerate downstream task performance for negligible cost: autonomous research saves time for a modest cost.

\paragraph{Agents need human ``babysitting'' to mitigate operational struggles.}
Autonomous research loops often require human input or interventions to resolve operational struggles.
All underlying LLM agents were susceptible to underutilizing GPU memory, often proposing a smaller batch size than what was able to be fit.
They also often went idle or stopped prematurely despite explicit guidelines to ``never stop''.

Each agent also exhibited distinct failure patterns:
Cursor Composer often re-executed duplicate experiments.
GPT-OSS failed to reliably spawn sub-agents, frequently explaining how to start the run instead of acting as the orchestrator in multi-agent frameworks.
Finally, Claude Sonnet sometimes opted to attempt to debug environment scripts rather than run experiments.

These operational struggles, which we partially address with additional harness components (Section \ref{sec:additional-harness}), are by no means comprehensive of areas where agents may require guidance.
Overall, autonomous research agents often suffer from operational struggles that require human ``babysitting''.

\section{Conclusion}
In this work, we perform a case study on telecom ticket retrieval to empirically investigate generalization of autonomous research to open-ended machine learning problems.
Based on our findings, autonomous research for open-ended problems shows promise, despite operational and epistemic limitations.
Combining the ability of agentic frameworks to systematically explore search spaces and predefined design choices with the intuition and scientific reasoning of human researchers is likely the sweet spot given today's tools and resources.

\section*{Limitations}
This study has several limitations. 
First, our work analyzes a case study incorporating a single industrial task, telecom ticket retrieval, and three selected LLM agents, and the extent to which our findings generalize to other research problems and agents remains unclear. 
The task provides a realistic setting with a well-defined downstream metric, but research problems with different objectives, datasets, or model architectures may exhibit different dynamics.
It is also possible that more recent, increasingly powerful agent systems do not align with our observations--they may not require human direction or operational harness to outperform human research output.

Second, the comparison between human-led and agentic research is not a controlled comparison. 
The autonomous campaigns start from a human-designed documentation, dataset, codebase, evaluation pipeline, and existing baselines.
And the human researcher, with access to internal knowledge and domain expertise, remains responsible for monitoring and stopping campaigns.
Thus, although we did our best to refrain from distilling internal knowledge and domain expertise to agents or our implemented harness, our results should not be interpreted as direct evidence that agents can independently reproduce the entire research process from scratch.

Third, our conclusions are based on a limited number of campaigns and experiments. 
In particular, the effects of agent choice, multi-agent organization, and informedness are difficult to isolate because these factors vary across campaigns and are not evaluated under a controlled experimental design. 
Similarly, our observations about agents favoring local optimization over higher-level architectural changes should be viewed as empirical observations from this case study rather than general properties of LLM-based research agents. 

Finally, our evaluation primarily measures downstream retrieval performance and campaign cost; it does not directly quantify the quality, novelty, or reproducibility of the entire scientific process.
Future work may benefit from broader evaluations across tasks and larger, controlled campaign sets that investigate beyond one-dimensional representations of research campaigns (i.e. a single retrieval performance).

\section*{Acknowledgments}
This work has been performed during the summer internship program at Nokia Bell Labs.
We thank Gabriel Górski, Jakub Kozerski, Bartlomiej Ruszaj, and Adrian Dudycz from Nokia Mobile Infrastructure for collecting the telecom-related troubleshooting tickets, fault analyses, technical analyses, and logs. We thank Ahmet Akyamac from Nokia Bell Labs for preparing the computational resources used to run our local open-weight models.

\appendix
\bibliography{aaai2027}
\section{Details on Best Discovered Model}
\label{sec:details-best-model}
We briefly describe the best discovered system from our autonomous research campaigns in Section \ref{sec:discovered_pipeline}.
Here, we describe the discovered representation and preprocessing strategy, training data generation, and details on architecture and hyperparameters that produce the best discovered system.
In addition to Recall@k metrics, we report $\text{MRR}@10 = \mathbf{0.4350}$, $\text{NDCG}@10 = \mathbf{0.2210}$, and $\text{MAP}@10 = \mathbf{0.1370}$.

\paragraph{Document Representation and Preprocessing.}
The three types of documents, \texttt{T}s, \texttt{FA}s, and \texttt{TA}s, are structured, JSON-style documents.
In our winning system, JSON-style documents are converted into a dense text input by concatenating the fields with explicit markers signaling the field.
This is comparable to BERT's \citep{devlin_bert_2019} traditional approach to natural language understanding tasks that use \texttt{[SEP]} tokens between sections of the input text.
In it, a two-stage text cleaning pipeline is employed.
First, empty fields are removed.
Then, a document-frequency line filter strips recurring boilerplate text appearing across 5 or more documents, removing system header noise that otherwise creates spurious lexical overlap. 

\paragraph{Training data generation.}
To compile a training dataset, we sample both positive pairs and mine negative ones from our dataset of \texttt{T}s, \texttt{FA}s, and \texttt{TA}s.
To capture structural relationships across documents, the pipeline constructs positive pair mappings within each cluster: \texttt{T}-to-\texttt{FA} ($270k$ base pairs) and \texttt{T}-to-\texttt{T} ($245k$ base pairs).
The winning model also configured a $2\times$ sampling weight to \texttt{T}-to-\texttt{T} pairs relative to \texttt{T}-to-\texttt{FA} pairs, expanding the effective dataset size to $760k$ training instances.
While the comparison systems in \citet{trabelsi_teledoctr_2026} rely on standard single-modality \texttt{T}-to-\texttt{FA} mappings, our discovered pipeline demonstrates that constructing and oversampling peer ticket relationships provides a more robust signal for document retrieval. 
For every positive pair, a single hard negative sample is mined.
The negative sample comes from outside the positive pair's cluster, but is still similar; it is sampled from \texttt{T}s with the same product and feature metadata information, two categorical fields in the \texttt{T} documents.
When no out-of-cluster \texttt{T} with the same product and feature metadata is available, negative sampling falls back to random cross-category sampling.

\paragraph{Model architecture and hyperparameters.}
The retriever utilizes a bi-encoder architecture initialized from a pre-trained MPNet base model \texttt{all-mpnet-base-v2} \citep{song_mpnet_2020}. Fine-tuning runs for 2 epochs using a batch size of 32, a learning rate of $\eta = 2 \times 10^{-5}$, and maximum length 384.
The resulting system's objective minimizes a Multiple Negatives Ranking Loss \citep[InfoNCE; ][]{oord_representation_2019} over cosine similarity with a fixed inverse temperature scaling factor of 20. 


\end{document}